%% file: paper.en.tex
\documentclass[11pt, a4paper]{article}

\input{macros.tex}

\hypersetup{
    colorlinks=true,
    linkcolor=blue,
    filecolor=magenta,      
    urlcolor=cyan,
    pdftitle={Compression is Routing},
    pdfauthor={Zhongpan Tang},
}

\title{\textbf{Compression is Routing: Reconstruction Error as an Intrinsic Signal for Modular Language Models}}

\author{Zhongpan Tang \\ 
        \texttt{tangzhongp@gmail.com}} 
\date{\today}

\begin{document}

\maketitle

\begin{abstract}
Current Large Language Models (LLMs) face three major challenges: context length limitations, high inference costs, and catastrophic forgetting during continual learning. While Mixture-of-Experts (MoE) architectures mitigate some of these conflicts, their routing mechanisms typically rely on explicitly trained auxiliary classifiers. This not only increases system complexity but also often lacks interpretability when handling mixed-domain inputs.

Building upon the premise that ``Compression is Intelligence,'' this paper proposes a novel architectural philosophy: \textbf{``Compression is Routing.''} We trained an 87M-parameter end-to-end Transformer Autoencoder, achieving a \textbf{64x sequence length compression} (compressing 512 tokens into 8 latent vectors). Experimental results demonstrate that this compressor possesses extreme domain discriminative capability: it achieves a reconstruction accuracy of \textbf{99.47\%} on the in-domain (code) validation set; accuracy drops sharply to \textbf{47.76\%} on a semi-out-of-distribution domain (Wiki text); and further plummets to just \textbf{0.57\%} on a fully out-of-distribution domain (random sequences).

This extreme and systematic performance discrepancy establishes the validity of reconstruction error as an \textbf{Intrinsic Distribution Fingerprint}. Based on this, we propose that expert modules can be automatically scheduled using reconstruction residuals directly, without the need for explicit gating networks. This mechanism offers excellent scalability. Furthermore, this architecture provides a new perspective on ``VRAM compression'' for handling ultra-long contexts. This report aims to verify the physical validity of this foundational architecture, offering a new research perspective for the next generation of scalable modular neural networks.
\end{abstract}

\section{Introduction}
\label{sec:introduction}

We firmly believe that \textbf{``Compression is Intelligence''} is the key principle guiding the evolution of language model architectures. Recent work by Delétang et al. (2024)~\cite{deletang_language_2024} rigorously demonstrated the equivalence between language modeling and lossless compression, pointing out that predictive capability and compression efficiency are essentially two sides of the same coin. However, while existing research strives to alleviate the long-context bottleneck through methods like Gist Tokens~\cite{mu_learning_2024}, ICAE~\cite{ge_-context_2024}, or CompLLM~\cite{berton_compllm_2025}, translating the theoretical limits of information theory into a deployable, high-density architectural paradigm remains challenging.

Existing compression pathways generally suffer from significant limitations. First is \textbf{suboptimal representation learning}: for instance, ICAE employs a frozen decoder strategy, forcing the encoder to unilaterally adapt to the original semantic space. This lack of end-to-end synergy limits the compression ratio (only about 4x) and hinders the establishment of efficient communication protocols. Second is \textbf{inference feasibility}: while Kuratov et al. (2025)~\cite{kuratov_cramming_2025} demonstrated a theoretical limit where a single vector can carry 1568 tokens, their approach relies on expensive Per-sample Optimization, lacking a generalized encoder, and the latent space is scattered, unsupported for downstream semantic operations. Finally, there are \textbf{physical bottlenecks of modalities}: the optical compression path attempted by DeepSeek-OCR~\cite{wei_deepseek-ocr_2025}, while innovative, is limited by image resolution and introduces cross-modal rendering overheads.

Meanwhile, at the macro-architectural level, as model scales grow, ``Monolithic Generalist'' architectures are facing the dilemma of diminishing marginal returns due to parameter interference and high continuous training costs. Modular design is an inevitable trend, but traditional Mixture-of-Experts (MoE) models rely on explicitly trained gating networks, which increase system complexity and lack interpretability in their decision logic.

Addressing these limitations, this paper proposes a new architectural paradigm: \textbf{``Compression is Routing.''} We constructed an end-to-end Transformer Autoencoder. By jointly training the encoder and decoder, we established a generalizable, highly efficient compression protocol, achieving \textbf{64x} efficient compression in the \textbf{Latent Space} rather than bit-stream space.

More importantly, experiments show that this end-to-end learning forces data to form clear \textbf{Manifold Separation} in the latent space. Based on this, we propose the core hypothesis: \textbf{The reconstruction quality (Reconstruction Loss) of a compressor dedicated to a specific domain can serve directly as a routing signal.} This discovery not only significantly alleviates VRAM bottlenecks in ultra-long context scenarios but also provides a new information-theoretic perspective for building modular neural network systems that do not require explicit gating parameters. By explicitly modeling reconstruction residuals as a measure of information deficiency, dynamic scheduling of expert modules is achieved. Thus, the model paradigm expands from ``Compression is Prediction'' to ``Compression is Routing.''

\section{Methodology}
\label{sec:model_and_method}

\subsection{End-to-End Language Autoencoder Architecture}
We adopt an asymmetric Transformer-based autoencoder architecture. The information flow is defined as:
\begin{equation}
x \to \text{Encoder} \to z \to [z, m] \to \text{Decoder} \to \hat{x}
\end{equation}
Where:
\begin{itemize}
    \item \textbf{Compressor (Encoder)}: Receives a token sequence $x$ of length $L=512$ and dimension $d_{model}=512$. Through attention mechanisms, it maps this to a latent vector sequence $z$ of length $M=8$. This achieves a \textbf{64:1} sequence length compression ratio.
    \item \textbf{Reconstructor (Decoder)}: Conditioned on $z$ and an auxiliary signal $m$, it reconstructs the original sequence $\hat{x}$. The loss function uses the standard \textbf{Cross-Entropy Loss} between the predicted \textbf{Logits} (vocabulary probability distribution) and the true labels $x$.
\end{itemize}

\subsection{Physical Isolation and Information Bottleneck}
To achieve ``pure'' compression and prevent the model from learning identity mappings via shortcuts, we implemented strict \textbf{physical isolation}:
\begin{enumerate}
    \item \textbf{Input Isolation}: The auxiliary signal $m$ has the same length as $x$ but contains no content information associated with $x$.
    \item \textbf{Information Path}: The decoder is blocked from accessing the original input $x$ and can only acquire semantic information through the highly compressed $z$ vectors. This isolation ensures that $z$ becomes the sole information bottleneck, forcing the encoder to learn the projection of the data manifold rather than simple signal transmission.
\end{enumerate}

\section{Experiments and Analysis}
\label{sec:experiments_and_analysis}

\subsection{Experimental Setup}
We used the GPT-2 Tokenizer and validated our approach on an \textbf{87M parameter} model. The model was pre-trained on a code dataset (\texttt{codeparrot/codeparrot-clean}). The evaluation benchmarks include three gradients of distribution:
\begin{enumerate}
    \item \textbf{In-Domain (ID)}: The code repository was split into training and validation sets. Testing was performed on the validation set after training.
    \item \textbf{Semi-OOD}: \texttt{Salesforce/wikitext-103} (Natural Language), using its validation set for testing.
    \item \textbf{Full-OOD}: High-entropy random token sequences.
\end{enumerate}

\subsection{Evaluation Metrics}

To quantify the information fidelity of the autoencoder during compression, we define \textbf{Token-level Reconstruction Accuracy (TRA)}.

Given an original input sequence $x = \{x_1, x_2, \dots, x_L\}$ of length $L$ and the model's reconstructed output sequence $\hat{x} = \{\hat{x}_1, \hat{x}_2, \dots, \hat{x}_L\}$, TRA is defined as the proportion of tokens that match at corresponding positions:
\begin{equation}
\text{TRA}(x, \hat{x}) = \frac{1}{L} \sum_{t=1}^{L} \mathbb{I}(x_t = \hat{x}_t)
\end{equation}
where $\mathbb{I}(\cdot)$ is the indicator function, taking the value 1 when the condition is true, and 0 otherwise.

\vspace{0.5em}
\noindent \textbf{Example}: Suppose the input sequence is $[1, 2, 3, 4, 5]$ and the reconstructed sequence is $[1, 2, 6, 7, 5]$. The tokens at positions 1, 2, and 5 are predicted correctly (3 total), while positions 3 and 4 are incorrect. The reconstruction accuracy for this sample is $3/5 = 60\%$. This metric strictly measures the model's precision in preserving sequence structure and specific content, rather than vague semantic similarity.

\subsection{Core Finding: The Three-Step Decay of Reconstruction Accuracy}

The experimental results are shown in Table \ref{tab:results}.

\begin{table}[h]
\centering
\caption{Comparison of Reconstruction Accuracy Across Different Data Distributions}
\label{tab:results}
\resizebox{\textwidth}{!}{%
\begin{tabular}{@{}llcl@{}}
\toprule
\textbf{Data Distribution} & \textbf{Validation Tokens} & \textbf{Reconstruction Accuracy} & \textbf{Routing Signal Meaning} \\ \midrule
In-Domain (Code)         & 3,170,816              & \textbf{99.47\%}               & Strong Positive (Perfect Fit)    \\
Semi-OOD (Wiki)          & 320,512                & \textbf{47.76\%}               & Moderate Negative (Structural Bias)\\
Full-OOD (Random)        & 3,170,816              & \textbf{0.57\%}                & Extreme Negative (Total Rejection)  \\ \bottomrule
\end{tabular}
}
\end{table}

\subsection{Geometric Verification and Mechanism Explanation}
\begin{itemize}
    \item \textbf{High Precision ID Reconstruction (99.47\%)}: Proves that $z$ possesses extremely high channel capacity, capable of perfectly carrying the structure of the code manifold with 64x compression.
    \item \textbf{Moderate OOD Decay (47.76\%)}: This figure is not a random guess (which would be approx. $1/|\text{V}| \approx 0.002\%$). It reflects that the model utilized the \textbf{shared vocabulary} and \textbf{fundamental statistical features} between code and natural language. This represents the overlapping region between manifolds and is the ideal signal to trigger an \textbf{Incremental Expert}.
    \item \textbf{Extremely Low F-OOD Rejection (0.57\%)}: Proves the model did not degenerate into a signal transmission line; it possesses strong rejection capabilities against noise that does not conform to the training distribution.
\end{itemize}

\subsection{Geometric Topology of Latent Space}

\begin{figure}[htbp]
    \centering
    \begin{subfigure}{0.48\textwidth}
        \centering
        \includegraphics[width=\linewidth]{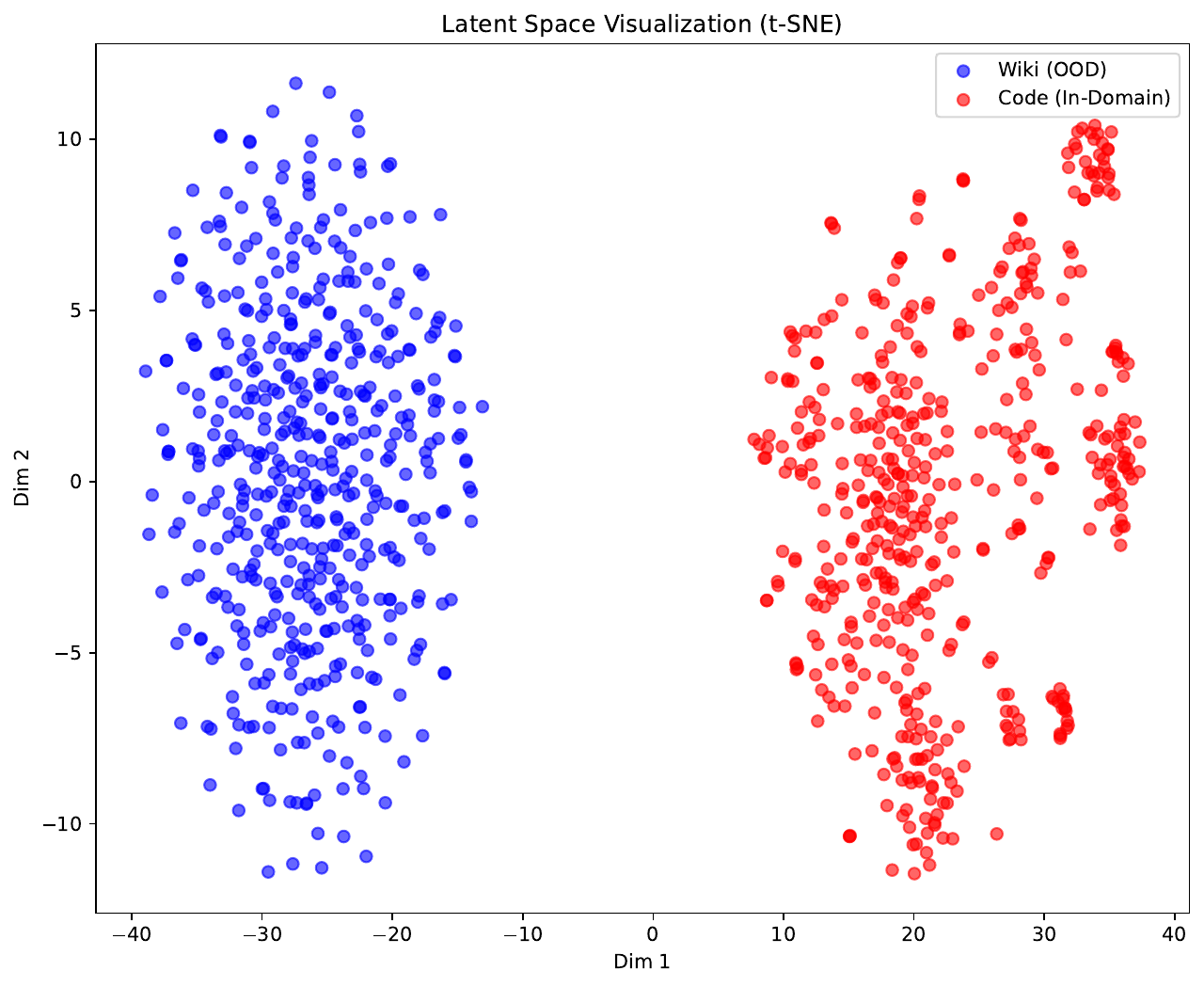} 
        \caption{t-SNE Latent Space Visualization}
        \label{fig:tsne}
    \end{subfigure}
    \hfill
    \begin{subfigure}{0.48\textwidth}
        \centering
        \includegraphics[width=\linewidth]{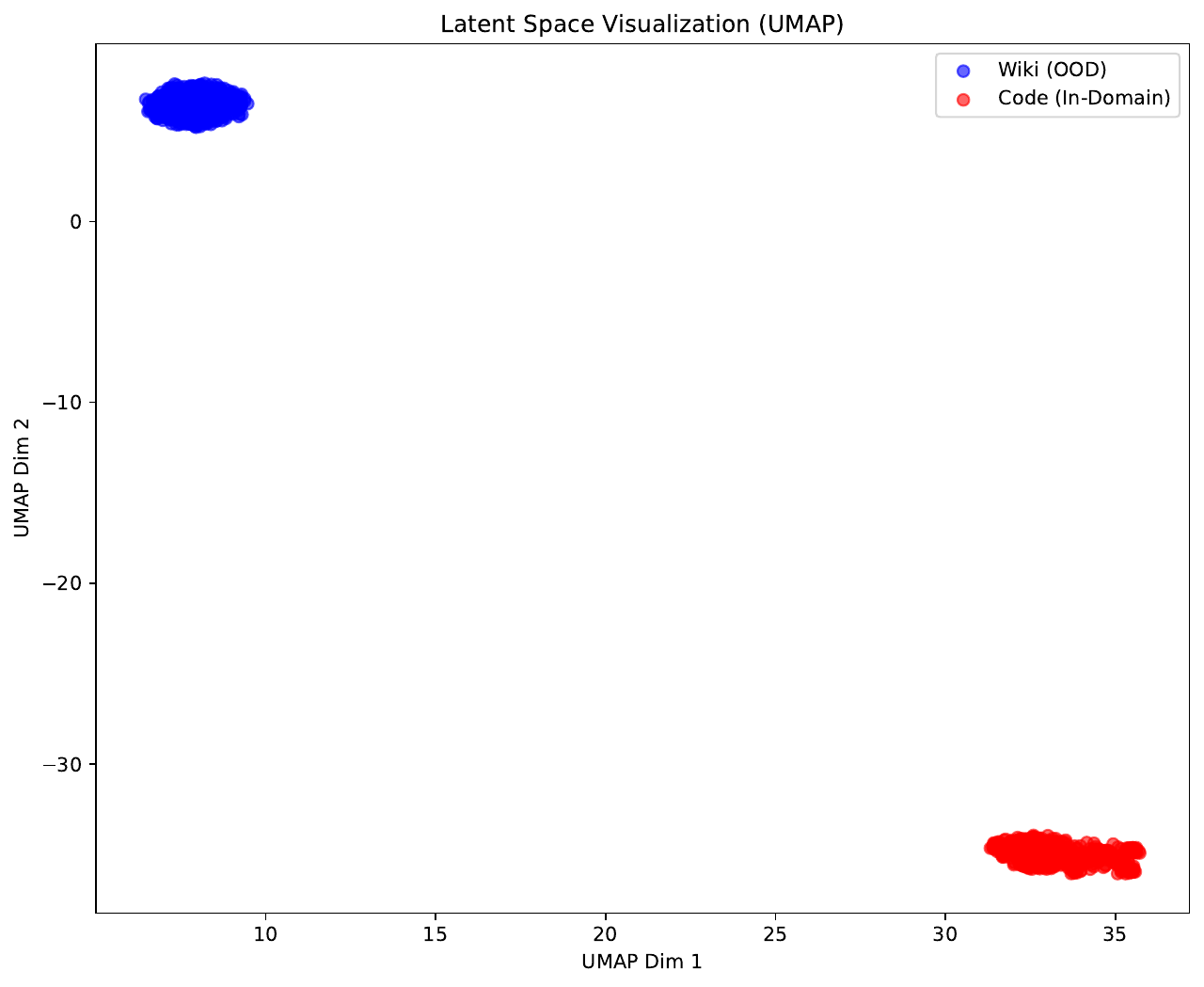}
        \caption{UMAP Latent Space Visualization}
        \label{fig:umap}
    \end{subfigure}
    \caption{\textbf{Visualization of Manifold Geometry in Latent Space}. Red points represent In-Domain data (Code), and blue points represent Out-of-Domain data (Wiki). Left: t-SNE shows local clustering characteristics. Right: UMAP shows global topological isolation. The two classes of data exhibit complete linear separability in the latent space.}
    \label{fig:manifold_vis}
\end{figure}

\begin{figure}[htbp]
    \centering
    \includegraphics[width=0.8\textwidth]{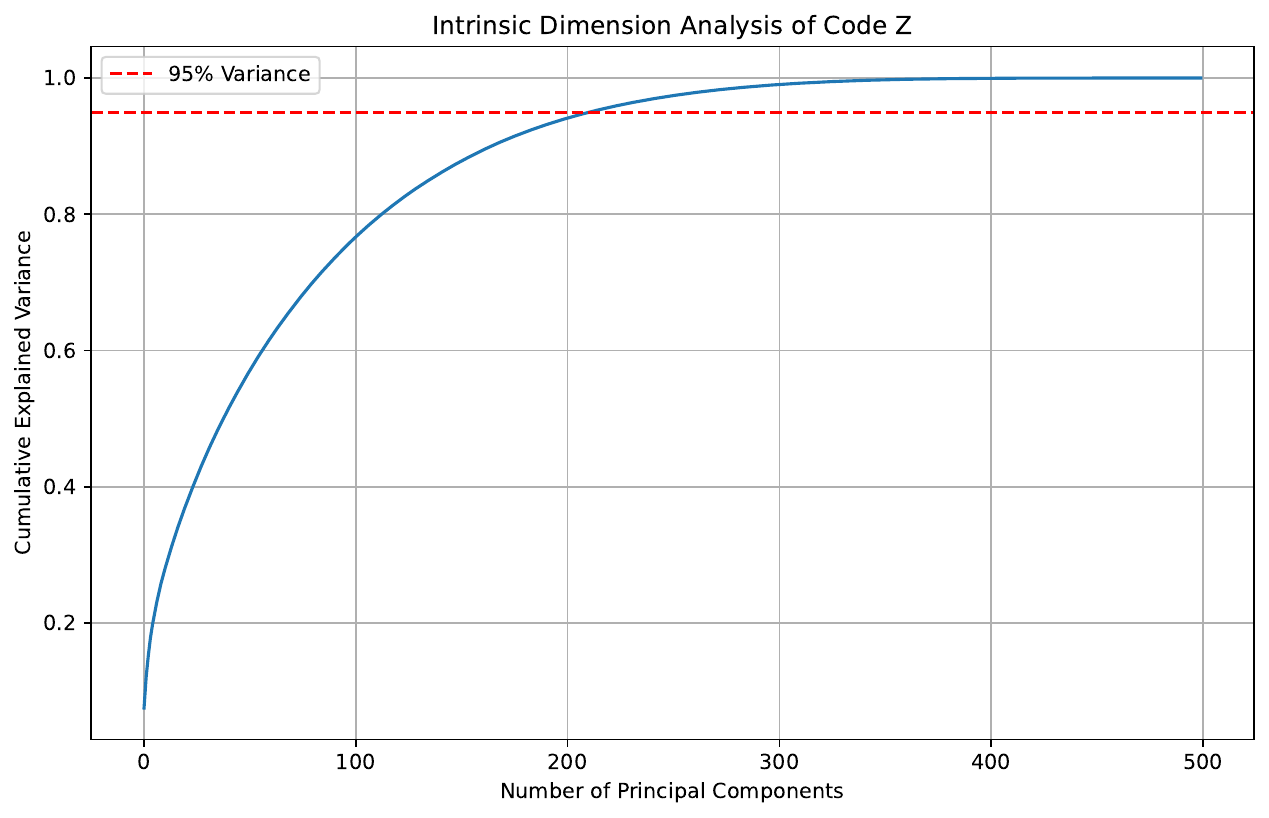}
    \caption{\textbf{Intrinsic Dimension Analysis of In-Domain (Code) Latent Vectors}. The curve shows the PCA cumulative explained variance ratio. The red dashed line marks the 95\% variance threshold.}
    \label{fig:pca}
\end{figure}

To investigate the physical mechanisms behind the reconstruction error differences, we utilized t-SNE, UMAP, and PCA techniques to visualize and analyze the dimensionality of the latent vectors $z$ generated by the encoder. Figures \ref{fig:manifold_vis} and \ref{fig:pca} reveal three key geometric properties:

\subsubsection{Global Orthogonality of Manifolds}
The UMAP visualization (Figure \ref{fig:umap}) reveals the most striking result: In-Domain data (Code, red) and Out-of-Domain data (Wiki, blue) appear as two extremely dense and disconnected islands in the global topology. There exists a vast vacuum between them, and their geometric distributions are nearly orthogonal.
This implies that the 64x compression bottleneck forces the encoder to project different data distributions onto mutually exclusive sub-manifolds in the latent space. This \textbf{Physical Isolation} provides the geometric foundation for alleviating catastrophic forgetting—training a Code expert will hardly interfere with the latent region occupied by a Wiki expert.

\subsubsection{Linear Separability for Routing}
The t-SNE visualization (Figure \ref{fig:tsne}) further confirms the robustness of this separation. Even from a perspective that preserves local neighborhood structures, the red and blue clusters remain separated by a clear boundary.
This geometric property proves that the \textbf{computational cost of routing decisions is virtually zero}. Since the two manifolds are highly linearly separable, the system does not need to train complex gating networks to achieve expert dispatching with extremely high accuracy.

\subsubsection{Sparsity and Low Intrinsic Dimension}
PCA analysis (Figure \ref{fig:pca}) reveals the efficiency of the encoder. Although the physical dimension of the latent vector $z$ is $D=512$, the cumulative variance curve shows that the first $\approx 200$ principal components are sufficient to explain 95\% of the data variance.
This indicates that the \textbf{Intrinsic Dimension} of the Code data is far lower than its physical dimension ($d_{intrinsic} \approx 200 < 512$). The encoder has learned to remove high-frequency noise and redundant information from the input, retaining only the most representative skeletal features.

\subsubsection{Supplementary Note}
While this report focuses on experimental results based on the standard Hugging Face ecosystem to ensure reproducibility and benchmark alignment for the community, it is worth noting that in the early exploration phase of this research, we conducted extensive tests based on \textbf{custom-trained tokenizers} and \textbf{real-world Chinese novel corpora}. All tests reproduced the exact same performance tiered phenomenon. This further confirms that the ``Compression is Routing'' mechanism does not rely on specific languages or tokenization strategies but possesses universal robustness across languages and systems.

\section{Discussion}
\label{sec:more_discussion}

\subsection{Iteration Cost and Catastrophic Forgetting: The Evolution Bottleneck of Monolithic Architectures}
The current LLM paradigm relies primarily on Joint Training over massive mixed data. While this ``Monolithic'' architecture performs well in static evaluations, its \textbf{Iteration Cost} has become an ignorable obstacle when facing dynamic knowledge environments.

\begin{itemize}
    \item \textbf{Prohibitive Re-training Costs}: In a monolithic architecture, injecting a new data distribution (e.g., a new programming language or private industry knowledge) often requires full fine-tuning or even re-training of hundreds of billions of parameters. This rigid update mode leads to astronomical computational consumption and immense time costs.
    \item \textbf{Catastrophic Forgetting}: Attempting to introduce a new domain into a monolithic model via fine-tuning can easily destroy the parameter distribution learned for old domains. To maintain general capabilities, massive amounts of old data must be mixed back in (Replay) during every update, further exacerbating training inefficiency.
    \item \textbf{Modular Solution}: In contrast, the ``Compression is Routing'' architecture offers a paradigm for \textbf{Incremental Expansion}. When facing a new distribution (e.g., Code), we do not need to touch existing expert modules (e.g., a Novel compressor). We simply mount and train a new residual compressor for that distribution. This \textbf{Parameter Freezing} strategy physically prevents interference with old knowledge, minimizing the marginal cost of learning new knowledge.
\end{itemize}

Therefore, we argue that future general intelligence systems should not be giant monoliths requiring repeated re-casting, but rather open ecosystems capable of \textbf{Continuous Evolution} through the low-cost mounting of new modules.

\subsection{Possibilities for Ultra-Long Context: The VRAM Revolution}
An unexpected but significant benefit of this architecture lies in its potential for handling ultra-long contexts.
In traditional Transformers, KV Cache VRAM usage grows linearly with sequence length. In our architecture, mapping representations after the embedding layer to a more compact latent space achieves a \textbf{64x Sequence Length Compression}.
\begin{itemize}
    \item \textbf{Compression Ratio}: Compressing 512 tokens into 8 vectors.
    \item \textbf{VRAM Advantage}: This means contexts that originally required multiple A100 GPUs can now be processed on a single consumer-grade graphics card.
\end{itemize}

\section{Limitations and Future Work}
\label{sec:limit_and_future_work}

\subsection{Limitations}
As a foundational architecture validation, the current work has limitations:
\begin{enumerate}
    \item \textbf{Hysteresis in Dynamic Switching}: The current architecture uses a fixed block length ($L=512$) for compression. This creates noticeable \textbf{Hysteresis} at boundaries where the distribution switches drastically. Since the latent vector $z$ encodes global distribution features of the entire block, when the input modality changes suddenly within a block (e.g., switching from literary narrative to a code block), the state of the $z$ space often reacts sluggishly due to being ``dominated'' by the preceding distribution, leading to abnormally high reconstruction errors at the switching moment.
    \item \textbf{Granularity Mismatch in Mixed Distributions}: Real-world data often manifests as \textbf{Interleaved Distributions} (e.g., technical blogs containing code snippets). Due to the fixed compression window, when a single slice contains data from different manifolds, a single expert struggles to accommodate both, causing reconstruction accuracy to fall into a ``gray area'' that is difficult to classify.
\end{enumerate}

\subsection{Resource Constraints and Researcher Statement}
As a project initiated and completed by an \textbf{Independent Researcher}, this study is subject to objective limitations in computational hardware and experimental cycles, which restricted our ability to conduct broader Ablation Studies. Specifically:

\begin{enumerate}
    \item \textbf{Minimum Convergence Length Observation}: In training for $L=512$ slices, we evaluated discrete samples of latent vector sequence lengths ($M \in \{2, 4, 8\}$). Results indicated that $M=8$ was the minimum length capable of achieving \textbf{Near-lossless Reconstruction} for this test set; in contrast, when $M=2$ or $M=4$, the model showed significant information loss and failed to effectively restore the input sequence.
    \item \textbf{Potential for Higher Compression Ratios}: Although this report mainly presents 64x compression results, early explorations suggested the potential to push the compression ratio to 128x or even higher. We hypothesize the existence of an information-entropy-based \textbf{Empirical Formula} or Scaling Law between latent vector length $M$ and sequence length $L$.
\end{enumerate}

The intention behind writing this technical report is to \textbf{inviting community exploration}: We hope that by demonstrating the physical validity of this core architecture, we can inspire communities or laboratories with richer computational resources to further explore the physical limits of encoders as knowledge containers.

\section{Conclusion}
\label{sec:conclusion}

The \textbf{``Compression is Routing''} paradigm proposed in this paper experimentally reveals the natural advantage of reconstruction accuracy as a routing signal. The significant gap of \textbf{99.47\% vs. 47.76\% vs. 0.57\%} establishes the legitimacy of reconstruction error as a metric for distribution determination.

This discovery addresses the issue of interpretability in MoE routing and provides a solid empirical foundation for building modular language model systems that are scalable, require no explicit gating, and possess extremely high inference efficiency (64x compression). We hope this technical report serves as a starting point to inspire deeper thinking within the community regarding the relationship between encoder performance and routing mechanisms.

\bibliographystyle{plain}
\bibliography{references} 

\end{document}

%% file: macros.tex
\usepackage[utf8]{inputenc} 
\usepackage{amssymb}        
\usepackage{float}          
\usepackage[margin=1in]{geometry} 
\usepackage{url}            
\usepackage{lmodern}        
\usepackage{tikz}
\usepackage{indentfirst}    
\usepackage{makecell}

\usepackage{siunitx}  
\usepackage{multirow} 
\usepackage{bm}

\usepackage{geometry}       
\usepackage{amsmath}        
\usepackage{amsfonts}       
\usepackage{graphicx}       
\usepackage{booktabs}       
\usepackage{hyperref}       
\usepackage{xcolor}         
\usepackage{fancyhdr}       
\usepackage{subcaption}     
\usepackage{abstract}       

\usepackage{sansmath}

\usetikzlibrary{
    arrows.meta, 
    decorations.pathreplacing,
    decorations.pathmorphing,
    positioning,
    calc,
    shapes.misc,
    fit,
    shadows,
    patterns,
    shapes.geometric,
    shadows.blur
}

\newcommand{\genCausalFullconnect}[6]{
    \foreach \i in {1,...,#3} {
        \foreach \j in {1,...,#6} {
            \pgfmathparse{ifthenelse(\i<=\j, 1, 0)}
            
            \ifnum\pgfmathresult=1
                \pgfmathsetmacro{\startx}{int(\i + #2 - 1)} %
                \pgfmathsetmacro{\starte}{int(\j + #5 - 1)} %
                \draw[->, >=latex] (#1-\startx) -- (#4-\starte);
            \fi
        }
    }
}